\pdfoutput=1

\documentclass[11pt]{article}

\usepackage[preprint]{acl}

\usepackage{times}
\usepackage{latexsym}

\usepackage[T2A,T1]{fontenc}

\usepackage[utf8]{inputenc}

\usepackage{microtype}

\usepackage{inconsolata}

\usepackage{graphicx}
\usepackage[russian,english]{babel}
\usepackage{tabularx}
\usepackage{enumitem}
\usepackage{booktabs}
\usepackage{amsmath}
\usepackage[most]{tcolorbox}

\newcolumntype{Y}{>{\centering\arraybackslash}X}

\title{Uncovering Differences in Persuasive Language in Russian versus English Wikipedia}

\author{Bryan Li \\
  University of Pennsylvania \\
  Philadelphia, PA, USA \\
  \texttt{bryanli@seas.upenn.edu} \\\And
  Aleksey Panasyuk \\
  Air Force Research Lab \\
  Rome, NY, USA \\
  \texttt{aleksey.panasyuk@us.af.mil} \\ \And
  Chris Callison-Burch \\
  University of Pennsylvania \\
  Philadelphia, PA, USA \\
  \texttt{ccb@cis.upenn.edu} \\}

\begin{document}
\maketitle
\begin{abstract}
We study how differences in persuasive language across Wikipedia articles, written in either English and Russian, can uncover each culture's distinct perspective on different subjects.
We develop a large language model (LLM) powered system to identify instances of persuasive language in multilingual texts. Instead of directly prompting LLMs to detect persuasion, which is subjective and difficult, we propose to reframe the task to instead ask \textit{high-level questions} (HLQs) which capture different persuasive aspects. Importantly, these HLQs are authored by LLMs themselves. LLMs over-generate a large set of HLQs, which are subsequently filtered to a small set aligned with human labels for the original task. We then apply our approach to a large-scale, bilingual dataset of Wikipedia articles (88K total), using a two-stage \textit{identify-then-extract} prompting strategy to find instances of persuasion. 

We quantify the amount of persuasion per article, and explore the differences in persuasion through several experiments on the paired articles. Notably, we generate rankings of articles by persuasion in both languages. These rankings match our intuitions on the culturally-salient subjects; Russian Wikipedia highlights subjects on Ukraine, while English Wikipedia highlights the Middle East. Grouping subjects into larger topics, we find politically-related events contain more persuasion than others. We further demonstrate that HLQs obtain similar performance when posed in either English or Russian.
Our methodology enables cross-lingual, cross-cultural understanding at scale, and we release our code, prompts, and data.\footnote{\url{https://github.com/apanasyu/UNCOVER_SPIE}}
\end{abstract}

\section{Introduction}
\begin{figure} [t!]
    \centering
   \includegraphics[width=\linewidth]{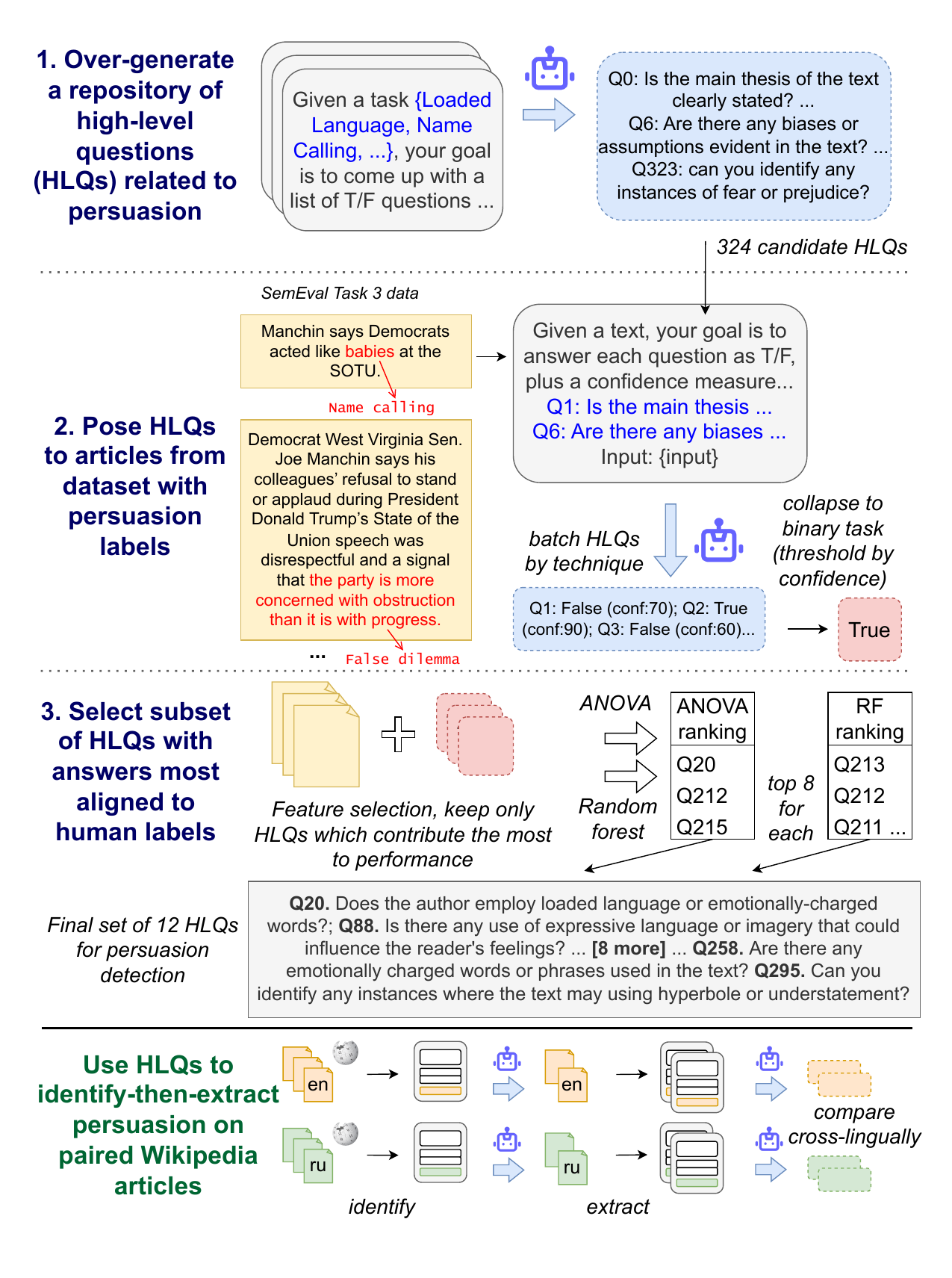}
   \caption{Overview of our approach for persuasion detection. \textbf{Top}: an LLM generates many high-level questions (HLQs), based on its own understanding of persuasion techniques. We then pose these HLQs to articles from an labeled persuasion dataset~\citep{piskorski2023semeval}, then select a subset of 12 questions which are most aligned to the human labels. \textbf{Bottom}: on another dataset, we use HLQs to prompt an LLM to \textit{identify-then-extract} persuasive spans. This is done over paired Wikipedia articles in Russian and English, facilitating cross-lingual comparison. }
   \label{fig:hlq_gen}
\end{figure} 

Wikipedia is a widely-used and comprehensive online encyclopedia. It is available in multiple languages, and as such, is accessed and trusted by users from countries and cultures across the world. On the same subject, different language Wikipedia articles are typically independently authored from one another -- although often with reference to the English version. Volunteer contributors follow a set of principles, among them to maintain a \textit{Neutral Point of View} (NPOV): that authors create and edit content with as objective of a view as possible.

Despite this guiding principle, Wikipedia has nevertheless come under criticism for perceived biases.
In fact, there is a Wikipedia article on ``Ideological bias on Wikipedia'' with ample discussion.\footnote{\url{https://en.wikipedia.org/wiki/Ideological_bias_on_Wikipedia}} The main concern is Wikipedia advancing liberal or left-leaning point-of-views. However, this is arguably a function of Wikipedia operating in a left-leaning news ecosystem, with one source opining ``The encyclopedia's reliance on outside sources, primarily newspapers, means it will be only as diverse as the rest of the media -- which is to say, not very''~\citep{kessenides2016woke}. These systemic biases arise, then, less as a conscious decision by authors, but more as a synthesis of the viewpoints from the primary sources.

This issue of authors' limited world-view compounds when considering authors who write in different languages. Russian and English Wikipedia articles often offer opposing views for many subjects. Authors in English will favor citations to English media, while authors in Russian have better access to Russian media. They also write in consideration of the interests and beliefs of their target audiences. Therefore, the same events and entities on Wikipedia have their content and tone shaped through language-specific cultural lenses. Even for the expressed goal of NPOVs, what is considered ``neutral'' can be subjective across cultures.

Prior work has either focused on English Wikipedia~\citep{hube2017bias,morris2023colonization}, or performed small-scale cross-lingual studies~\citep{zhou2016who, aleksandrova2019multilingual}.
In this work, we perform a large-scale study on how cross-cultural perspective differences manifest in Wikipedia.  We propose to quantify these articles' differences through identifying instances of \textit{persuasive language} -- how it is used, how much it is used, and for when it is used. 
We consider 26K Wikipedia subjects of interest to both cultures, and develop a large language model (LLM) powered system to automatically identify instances of persuasive language. 

Our approach is depicted in Figure~\ref{fig:hlq_gen}, and our contributions are:
\begin{enumerate}[noitemsep,topsep=0pt]
    \item  We develop an LLM-powered system to identify instances of persuasive language in English and Russian texts, which automates insights at scale.
    \item  We find that a baseline approach, which directly asks an LLM to identify persuasion used in a text,  results in responses that are over-sensitive and over-confident.
    \item We propose a novel framework of \textbf{high-level questioning}, which reframes the persuasion detection task into a set of high-level questions (HLQs). A large number of HLQs are LLM-authored, and are then filtered down to a small set best aligned to human labels of persuasion. On a binary persuasion detection task, HLQs achieve a 23.5\% relative improvement in F1 (.751 > .608).
    \item We study a large-scale dataset of 88K Wikipedia articles, with articles paired by subject in Russian and English. We extract persuasion with an \textbf{identify-then-extract} prompting approach with HLQs, reducing inference costs by 85.2\%.
    \item We perform several experiments into Wikipedia's cross-cultural differences in perspective, with metrics to quantify the amount of persuasion within a text. Experiments include ranking subjects by their salience to each language, and comparing persuasion between paired articles.
\end{enumerate}

\section{Related Work}
\label{sec:related}
\paragraph{Biases in Wikipedia}
Because of community guidelines such as NPOV, explicit biased statements in Wikipedia articles are removed by editors. Therefore, biases occur more subtly, through being systemic or implicit. Implicit bias occurs when  articles selectively choose what details to emphasize or omit~\citep{hube2017bias}. Identifying implicit bias in one article thus requires reference to another. Several authors use temporal edits of Wikipedia as references~\citep{morris2023colonization, yasseri2014most}. They identify from the editing cycle which viewpoints are removed (biased against), and which are kept (biased towards).
Our work takes a cross-cultural perspective, instead of a temporal one, in identifying biases; one language's article use of persuasion is compared against another.

Other works have studied how Wikipedia can be biased across languages.  ~\cite{zhou2016who} study how sentiment differs towards \~{}200 entities in 5 languages. ~\cite{aleksandrova2019multilingual} develop a system to extract biased sentences in 3 languages. There are several other relevant studies~\citep{callahan2011cultural,miz2020trending}. Our work is characterized by its much larger-scale (26K subjects), and its approach to extract potential bias at the span-level.

\paragraph{Multilingual biases of LLMs}

While LLMs are able to understand and generate text in many languages, researchers have identified that LLM competency and responses differ cross-lingually. For cultural inquiries, LLMs favor Western values, even when interacting in languages where different cultural sensitivities are desired~\citep{naous2023having,cao2023assessing}. For factual inquiries, multilingual settings cause LLMs to answer inconsistently~\citep{li2023land,qi2023cross}.

\paragraph{Russian vs Western perspectives}
The Russian state has positioned itself in stark contrast to the West. As such, Russia has made concerted efforts to spread its narratives and alter public discourse in its favor. This ranges from foreign events to domestic issues: respectively, the 2016 US Presidential Election~\citep{golovchenko2020cross}, and the 2022 Russian invasion of Ukraine ~\citep{geissler2023russian}.

Our work also seeks to compare Russian vs. Western POVs, but through comparing Russian articles written for Russian audiences, to English articles written for English audiences -- both of which aim for ``neutral'' POVs.

\paragraph{AI-assisted report generation}
This line of work uses AI tools to take in multiple documents, and assemble a report which summarizes the key points for a specified audience.  \citet{barham2023megawika} consider Wikipedias in 50 languages, to generate a large-scale dataset of 120m QA pairs, indexed to 71m reports. For a given passage, its citations are used as reports, and an English question-answer pair is generated from it. \citet{Li2023PAXQAGC} propose a scalable approach to generate cross-lingual QA pairs on paired passages. \citet{reddy2023smartbook} develop a LLM-powered system to generate reports to assist decision-makers in high-stakes issues. Our work takes inspiration from all of these, in studying Wikipedia, making cross-lingual comparisons, empowered by LLMs.

\section{Task Formulation}
\label{sec:problem_setup}

\paragraph{Definitions Used}
We tackle \textbf{detection of persuasion} in text.
We adopt two task formulations from the SemEval series of workshops. The first and simpler task is \textit{identification} -- predict whether a given context contains persuasion. The second is \textit{span extraction} -- given a context, extract spans that utilize persuasion. For either task, \textit{classification} can either be binary, or on a set of persuasive techniques (described in \S\ref{sec:dataset}). 

We consider two languages in this work, Russian and English. Our prompting setups are \textit{multilingually monolingual}, in that we cover both languages individually; i.e., for Russian contexts we use Russian prompts (and vice versa for English). Our analysis, however, will be \textit{cross-lingual}, in that we compare persuasion use across the paired articles, as well as compare across the entire Russian dataset vs the entire English dataset.

\textbf{Prompting} is the paradigm of interacting with LLMs at inference-time by giving instructions (a prompt) for a specific task. To further improve LLM's understanding of the task, \textit{few-shot} examples of the expected input and output can be added to the prompt. This is called \textit{in-context learning}~\citep{brown2020language,patelbidirectional}.

In this work, we consider a standard prompting setup with chat-optimized LLMs. Instructions are in the \textit{system prompt}, and the few-shot exemplars\footnote{We used a 1-shot, static exemplar for every prompt.} are captured in alternate \textit{user} and \textit{agent} sections. One inference entry is given as another user section, and the LLM will generate text to complete the agent section.

\subsection{Datasets Used}
\label{sec:dataset}

For designing and validating our prompts, we use SemEval 2023 Task 3 subtask 3~\citep{piskorski2023semeval}. We will simply refer to this as SemEval. The dataset covers 9 languages. Each article is segmented into paragraphs, and human annotators extract spans with persuasion, and also assign one of 23 persuasive techniques. Though we piloted some multilingual experimentation, we primarily work with the English subset of 11,780 paragraphs.

\paragraph{Selecting a Dataset in Russian and English}
We collect a set of paired Wikipedia articles, between Russian (ru) and English (en). We download the full dumps of Wikipedia in both languages, then filter to the subjects where articles link to known Russian state-sponsored news websites. 

In addition to the ru and en settings, we consider 2 more: English translated\footnote{This was paragraph-level MT by prompting an LLM. The full texts for all prompts used in this work are Appendix~\ref{sec:prompts}.} to Russian (en2ru), and the Russian translated to English (ru2en).

The final dataset consists of 22,046 paired articles; given the 4 settings, we will process 88k individual articles. At the paragraph-level, there are 245,778 ru entries (and ru2en), and 295,158 en entries (and en2ru), for >1m entries total.

\section{Baseline for Persuasion Detection}
\label{sec:identify}

\begin{figure} [t!]
    \centering
   \includegraphics[width=\linewidth]{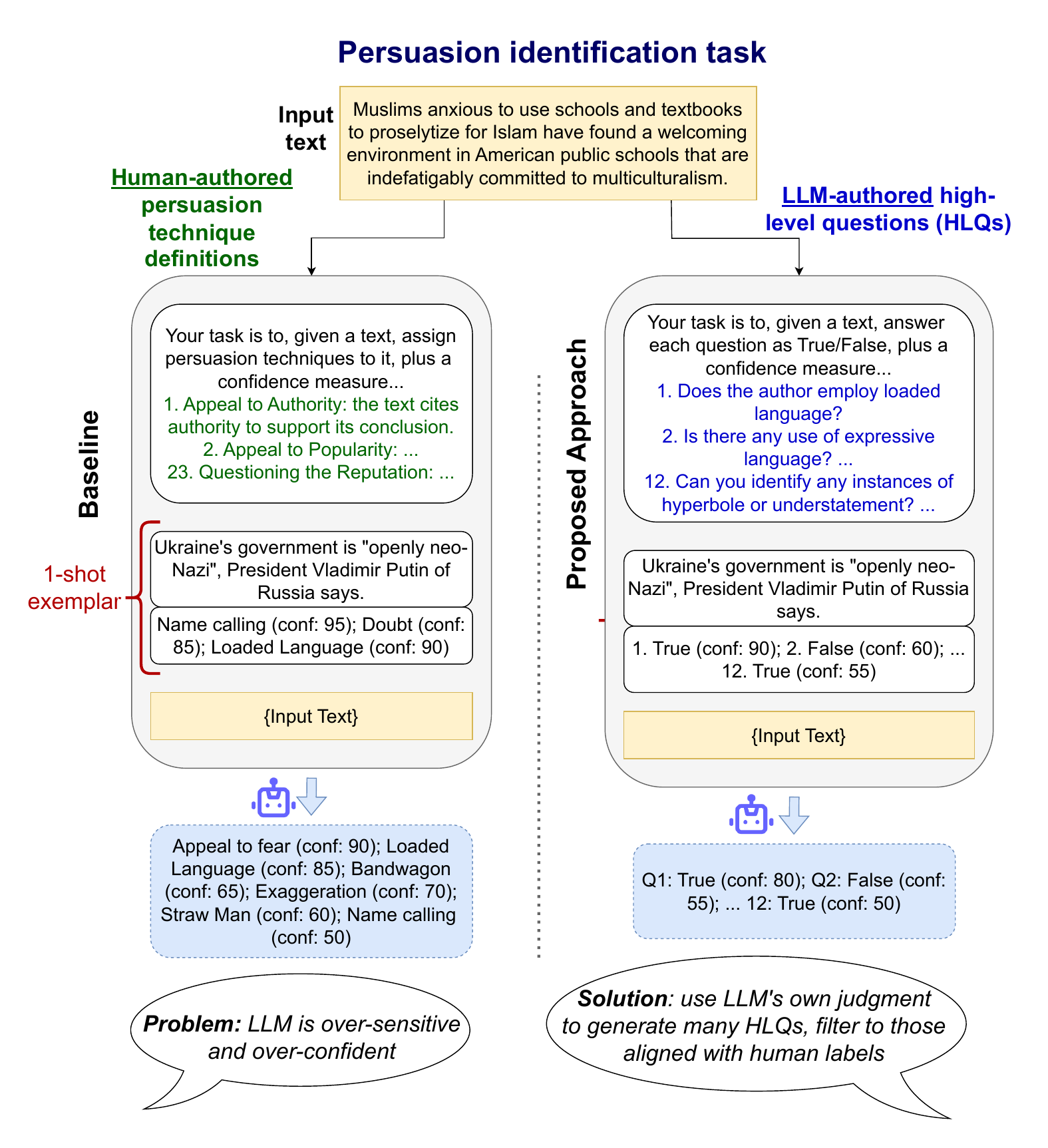}
   \caption{A comparison between two prompting approaches to persuasion technique detection. The baseline (left) directly uses the human-authored definitions. However, as these definitions were written for trained human annotators, the LLM misunderstands them and is over-sensitive and over-confident. Our proposed approach (right) instead leverages the LLM to decompose the task itself. Specifically, we elicit HLQs with a separate prompt (see Figure~\ref{fig:hlq_gen}). Then, we prompt with HLQs instead of definitions.}
   \label{fig:compare_prompts}
\end{figure} 

Figure~\ref{fig:compare_prompts} provides a comparison of the two approaches towards identifying
whether a context contains persuasion. In this section, we detail the baseline (left), which suffers from too many false positives. For this stage, we consider only English; we consider both languages in future sections.

\subsection{Approach: Direct Prompting with Definitions}
\label{sec:baseline}

The baseline prompt, as shown in Figure~\ref{fig:compare_prompts} (left) includes each persuasive technique, as well as a human-authored definition from SemEval. We further ask the LLM to generate a confidence score for each predicted technique, which we use to threshold predictions (described shortly ahead).

The main issue with this direct approach is that understanding of persuasion, is extremely subjective. In collecting the gold labels for SemEval, \citet{piskorski2023semeval} invested significant efforts into training 35+ human annotators (multilingually), and revising instructions throughout. So, by directly giving the LLM the final definitions, we cannot expect it to be aligned with the judgements specific to this annotation task.

To demonstrate the divergence in LLM understanding of persuasion vs. humans, Appendix Table~\ref{tab:pt_counts} compares the raw counts for each persuasion technique over SemEval (11,780 total contexts).  These numbers use the best observed confidence threshold of $x\geq 50$.
We observe that the gold labels are highly imbalanced. Notably, 59\% of texts contain no persuasion (6,945); however, GPT-4 predicts  ``no persuasion" only 12.3\% of the time (1,450). 
For the gold labels, 47\% of the persuasive texts receive the \textit{Loaded Language} label. 11 classes appear less than 1\% of the time. Furthermore, as this is a multi-class labeling task, those <1\% labels often appear with `Loaded Language'.
GPT-4 does not have a sense as to the class priors, and over-predicts the prevalence of all 23 persuasion techniques -- 24,209 predicted vs. 7,465 gold. 

For example, consider \textit{Appeal to Authority} (6286 vs. 179). The baseline prompt incorrectly assigns this to most mentions of people's titles (e.g. ``President Vladimir Putin of Russia'') or news sources (e.g. ``New York Times'').

\paragraph{Baseline makes LLMs over-confident}
To evaluate how confidence scores affect performance, we make a task simplification -- rather than multi-class labeling, we reduce the problem to binary classification. A context is `True' if any predicted technique has confidence $\geq x$, else `False'. We then use F1 to choose the optimal threshold.

\begin{table}[t!]
\centering
\small
\caption{F1 on SemEval binary persuasion detection, using the Baseline prompt, at varying confidence thresholds. Observe that \# `True' has too many false positives at low thresholds.}
\label{tab:conf}
\begin{tabular}{@{}lll|lll@{}}
\toprule
 $\geq x$ & F1 & \# `True' &  $\geq x$ & F1 & \# `True' \\ \midrule
$x=20$ & 0.469 & 10079 & $x=60$ & 0.450 & 10266 \\
$x=30$ & 0.459 & 10165 & $x=80$ & 0.582 & 7233 \\
$x=40$ & 0.454 & 10223 & $x=85$ & \textbf{0.608} & \textbf{4264} \\
$x=50$ & 0.447 & 10293 &  $x=90$ & 0.573 & 2833 \\
\bottomrule
\end{tabular}
\end{table}
Table~\ref{tab:conf} shows F1 by confidence threshold. Until $x=60$, the model assigns $>87$\% of texts as containing persuasion. F1 is maximized at $x=85$, at 0.608. We thus have shown both that the model is over-confident, and that thresholding for confidence scores substantially improves performance. 

\section{High-Level Questioning (HLQ)}
\label{sec:hlq}

\begin{table}[t!]
\centering
\small
\begin{tabular}{cccc}
\toprule
\textbf{Method} & \textbf{P} & \textbf{R} & \textbf{F1} \\ \midrule
24 definitions &  0.607 & 0.613	& 0.608 \\
12 HLQs &  0.757 & 0.748 & 0.751 \\ 
324 HLQs&  0.746 & 0.733 & 0.737 \\ 
\bottomrule
\end{tabular}
\caption{SemEval performance with different methods.} 
\label{tab:semeval_perf}
\end{table}

The high-level questioning approach to persuasion detection is depicted at the top of Figure~\ref{fig:hlq_gen}. 
The idea behind HLQs is to leverage LLMs' own (many different) intuitions on a task, then filter down to those that best align with gold labels on a reference dataset. Appendix~\ref{sec:why_high} details the motivation behind HLQs. In this section, we describe the approach the persuasion detection using HLQs.

\subsection{Generating candidate questions for each persuasion technique}
In the first step, we write a simple zero-shot prompt which tasks an LLM to generate list of True/False questions for a specified persuasion technique. For this step, the key is getting LLM's zero-shot understanding from various angles through its own generations. Therefore, we over-generate a large set of questions. It is expected that many questions overlap in coverage; we thus filter out questions which have very high n-gram overlap, while keeping paraphrases. This results in a repository of 324 questions.
Our manual analysis finds that most questions are very targeted, thus less subjective to answer (examples of questions in Table~\ref{tab:hlqs}).

\subsection{Applying HLQs to a labeled dataset}
Given the repository of HLQs, can we find which ones are most effective at detecting persuasive language? We do so by leveraging existing annotations from SemEval for the ground truth (step 2 of Figure~\ref{fig:hlq_gen}).
We batch the queries with sets containing all generated HLQs for a technique. Then, in a single prompt, an LLM is asked to answer True/False for the batched HLQs over the entire SemEval dataset (11,780 entries).

To compare to the gold annotations, we follow Section~\ref{sec:baseline} to
simplify and collapse SemEval to a binary classification task. As shown in Table~\ref{tab:semeval_perf}, prompting with HLQs improves F1 by 23.5\% relative over the baseline: F1 of 0.751 > 0.608. Furthermore, we see that the 12 question subset slightly improves over the full set of 324 HLQs: 0.751 > 0.737.
This shows using the LLM's own generations greatly improves over using definitions.

\subsection{Selecting subset of most-aligned HLQs}
The approach so far works well, but is expensive, with one multi-question prompt for each of the 23 techniques. We improve prompting efficiency by filtering to a subset of top-ranked HLQs, which maintains performance, while fitting into 1 prompt (step 3 of Figure~\ref{fig:hlq_gen}).

We cast this as a feature selection problem, which can be solved with the standard techniques of ANOVA and Random Forest with Gini impurity. 
Appendix Figure \ref{fig:feature_selection} illustrates the impact of feature reduction on the classifiers' effectiveness by plotting the F1-score against the progressively diminished feature sets using ANOVA. We find a stable performance across classifiers until around 8 features remain.

Thus, we combine the top 8 features from ANOVA, and top 8 from Random Forest. This results in a final subset of 12 HLQs. Appendix Table~\ref{tab:hlqs} shows both English and Russian versions.

\paragraph{Extending HLQs to Russian}
With the top 12 HLQs selected, we employ a native Russian speaker (one of the co-authors) for translation. They were allowed to prompt GPT-4, for assistance, before further postediting.\footnote{We follow the same LLM + human post-editing process to translate all prompts.}

\section{Identify-then-Extract Methodology}
\label{sec:methodology}
We adopt a two-stage hierarchical prompting approach   
towards persuasive language detection, using GPT-4\footnote{We also tried Llama-2, an open-source, much smaller LLM. With some prompt engineering, Llama-2 could do the task, though underperforming GPT-4 (see Appendix~\ref{sec:llama}).}, which we term identify-then-extract (Figure~\ref{fig:hlq_gen}, bottom). 

We apply identify-then-extract to find persuasion in the dataset of paired Wikipedia articles. Wikipedia is of particular interest because its use of persuasion tends to be more subtle, given that news articles often intend to tell a story, Wikipedia articles are all written to maintain NPOV.

Identify-then-extract thus is a further decomposition of the persuasion detection task, beyond the HLQ decomposition. Importantly, the identify step allows better identification of texts that are `Null' for persuasion (more common due to NPOV), and is much more efficient in terms of number of prompts, as described ahead.


\subsection{Identify}
Identification is the same task performed in \S\ref{sec:baseline}. In this stage, for each context, we prompt an LLM to answer True/False for all HLQs at once (shown in Figure~\ref{fig:compare_prompts}, right). This results in judgments for 12m (1m paragraphs * 12 HLQs) entries.

\subsection{Extract}
Of the 12m judgments, we only consider the contexts and the selected set of HLQs marked as `True'. For the paired Wikipedia articles dataset, 85.2\% are marked `False', and so do not need to be queried -- this shows the identify-then-extract approach saves much inference costs over a single-stage.

For each, we insert the context and one HLQ into a prompt template, which tasks the LLM to extract spans employing that HLQ. In contrast to the single prompt per context from the identify stage, the extract stage is hierarchical, having a set of `True' HLQs, and thus prompts, per context.


\paragraph{Collapsing extracted spans which overlap}
The HLQs, while nuanced, largely cover the same aspects of persuasion. This means that LLM outputs will also contain many overlapping terms. Given that for analysis purposes, we reduced the task from multi-class labeling to binary labeling, we should also collapse the multi-class extracted spans to a deduplicated set, termed a \textbf{persuasive text set (PTS)}. Appendix Table~\ref{tab:questionResponseExample} shows some sample model responses and the PTS.

\section{Experiments and Analysis}
\label{sec:analysis}
For our cross-lingual analysis over the paired articles dataset, we propose several metrics. We use these for various experiments, which make different comparisons and aggregations.

We note that these experiments proceed on a dataset which is unlabeled for persuasion. This is by design, as we would like to use the insights (i.e,. HLQs) generated from the limited amount of labeled data for a different domain, SemEval, and apply it to this huge dataset of 88K articles.\footnote{We performed manual analysis of a few examples from both languages. We acknowledge that followup work should take a closer look at how Wikipedia texts use persuasion.}

\paragraph{Metrics to Quantify Persuasion}
We define several metrics. $wc(\text{text})$ counts the number of words in a text.\footnote{We use the function \texttt{nltk.tokenize.word\_tokenize}.} The metrics are Persuasive Count (PC), and Persuasive Frequency (PF):
\begin{align*}
    PC = wc(\text{\text{PTS}})  \quad ; \quad
    PF_{\text{para}} = \frac{wc(\text{PTS})}{wc(\text{para})} \\
    PF_{\text{article}} = \sum_{\text{para} \in \text{article}} PF_{\text{para}} * \frac{wc(\text{para})}{wc(\text{article})}
\end{align*}

Our quantification of persuasion is more fine-grained than as done by prior works such as SemEval, which counts spans, rather individual words.\footnote{We acknowledge that word counting is simple, and that future work should precisely explore persuasion metrics.} 
We consider the persuasive text sets obtained with the identify-the-extract with HLQs approach. We find that over the 22K Russian articles, PF $\mu=.116, \delta=.177$, and for the English articles, PF $\mu=.137, \delta=.186$.

We next describe the experiments: a targeted case study, ranking articles by persuasion per-language, and several cross-lingual experiments.

\subsection{Case study: 2021 Russian protests}
\label{sec:case_study}

\begin{figure*}[t!]
    \centering
   \includegraphics[width=.97\linewidth]{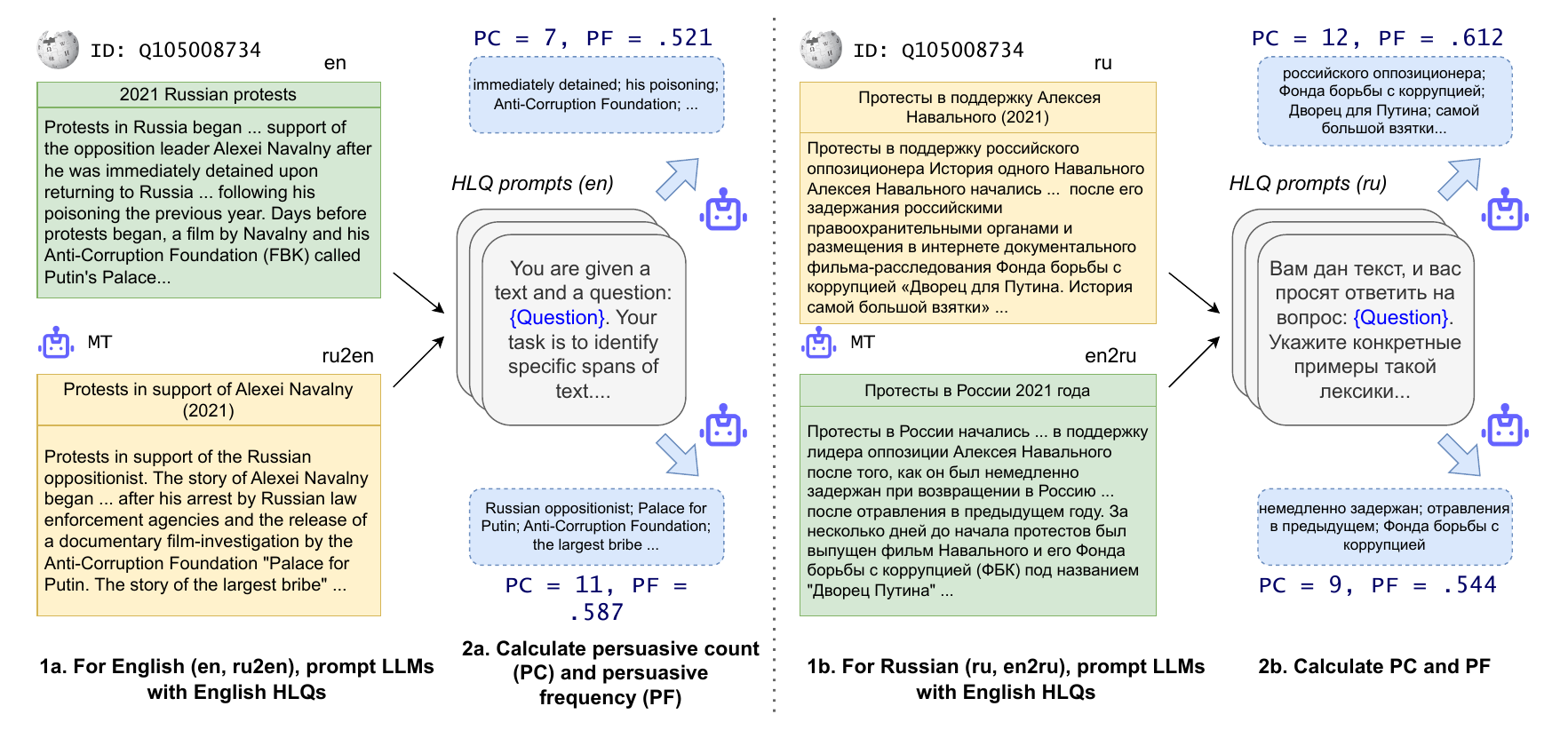}
   \caption{Depiction of the method to compare persuasive language usage across languages. For each language, we use HLQ prompts \textit{monolingually} on all articles to extract persuasive text spans (left: en, right: ru). We compare both persuasive count (PC) and persuasive frequency (PF) between the paired articles. For this case study, the Russian article (and its translation ru2en) are more persuasive on `2021 Russian protests'.  }
   \label{fig:xling_ef}
\end{figure*} 

\begin{table}[t!]
\centering
\small
\begin{tabularx}{\linewidth}{Y|ll} 
\toprule 
\textbf{Top Russian Articles  } & \textbf{\underline{PF ru}} & \textbf{PF en} \\ \midrule
Environmental impact of the 2022 Russian invasion of Ukraine & 0.982 & \textcolor{lightgray}{0.987} \\ 
Russian occupation of Kherson Oblast & 0.953 & \textcolor{lightgray}{0.651} \\ 
Cult of personality & 0.913 & \textcolor{lightgray}{0.608} \\ 
Disinformation in the 2022 Russian invasion of Ukraine & 0.911 & 0.819 \\ 
Trumpism & 0.879 & 0.828 \\ \toprule
 \textbf{Top English Articles } & \textbf{PF ru} & \textbf{\underline{PF en}} \\ \midrule
Ruscism & 0.778 & 0.882  \\ 
2015–2016 wave of violence in the Israeli–Palestinian conflict & 0.448 & 0.863  \\
Armenian genocide denial & 0.618 & 0.857 \\
Transphobia & 0.494 & 0.839 \\
Trumpism & 0.879 & 0.828 \\
\bottomrule
\end{tabularx}
\caption{Top 5 Wikipedia articles per language, ranked by persuasion frequency (PF). Each per-language ranking considers only the top 25\% articles by length. The other language scores are provided for reference; numbers in \textcolor{lightgray}{grey} indicate that the other language's article was below the length threshold. }
\label{tab:featuresForRank}
\end{table}

Figure~\ref{fig:xling_ef}, depicts a case study on paired articles for the subject `2021 Russian Protests'. Interestingly, the paired articles have different titles, as the Russian one is more specific, saying the protests were in support of Alexei Navalny.
The LLM extracts more persuasion from the Russian-authored articles than the English-authored -- .521 en vs. .587 ru2en. It identifies the loaded term ``oppositionist'' in the ru2en article. Meanwhile as ``opposition leader'' in the en article is more neutral, the term is not identified.

We also see that PF are relatively closer between an article and its translation: .521 en vs .544 en2ru; .612 ru vs .587 ru2en. This is a positive signal that LLM extracts similarly whether using the Russian or English prompts.
We also enlisted a native Russian speaker to verified these translations had the similar meanings. This sanity check is expanded, and applied to a larger dataset in Section~\ref{sec:sanity}.

\subsection{Ranking Wikipedia Articles by Persuasion}
\label{sec:rank_persuasion}

This experiment investigates which subjects contain the most persuasive content, as measured by PF, for Wikipedia authors in either language. We heuristically consider only the top 25\% longest articles, using length as a proxy for which articles are of the most interest to readers and authors.\footnote{for en: $wc>4758$,  for ru: $wc>2931$}

Table \ref{tab:featuresForRank} shows the two rankings for the top-5 persuasive articles. First, considering Russian, we see that 3 of the 5 subjects all deal with the 2022 Russo-Ukrainian war. The English versions of these 3 articles are below the top-25\% threshold; still, considering their scores, we see that ``Environmental impact'' has high PF en, while the other 2 are lower.

We now consider the English rankings. The top 5 subjects are more generally scoped, but align well with our intuitions on subjects of greater interest to Western audiences: ``Ruscism'' (Russian fascism), ``Transphobia'', ``Israeli-Palestinian conflict'', ``Armenian genocide denial''. The PF ru scores for these articles are much lower than for the Russian ranking.
Interestingly, we see that ``Trumpism'' is in the top-5 for both rankings, showing this political philosophy is of great interest to both societies.

\subsection{Grouping into Broader Topics}
\label{sec:topics}

We further our analysis by grouping subjects from Wikipedia into broader topics. For this, we leverage the WikiData knowledge base (KB), which contains structured KB triplets for every Wikipedia entry.
Specifically, we consider the predicate \texttt{is instance of} (Wikidata ID P31). We also use a normalized version of PF (NPF) to better compare scores across languages; calculation of NPF is described in Appendix~\ref{sec:npf}.

The NPF rankings by topic for en and ru settings are shown in Appendix Table~\ref{tab:top5_P31_Nodes}. The top topics as expected, such as `Disagreement Situation', and `Part of War'. The bottom topics are also as expected, such as `Aircraft' and `Automaker'. 
We have therefore validated our hypothesis that political-related events contain more persuasive content in both languages. More neutral categories, meanwhile, are written by both Russian and English authors with less persuasion.

Furthermore, we see that NPF scores are fairly well-aligned across languages. This could be in part due to neutral POV, and/or from the normalization process. This is an interesting finding, which shows that, despite individual subjects differing levels of persuasion across languages (as found in Section~\ref{sec:rank_persuasion}), within aggregated topics they are similarly persuasive. 

\subsection{Identifying subjects with the greatest cross-cultural disagreement}
\label{sec:disagreement}

For certain subjects of national pride, one culture may perceive it to be especially sensitive, and thus use more persuasion, than the other culture. We identify these by finding the paired articles with the largest PF differences.

Appendix Figure~\ref{fig:disagree} depicts selected subjects in a scatterplot, and again brings up interesting insights. We consider several examples and provide some cursory analysis and discussion.
The `1998 bombing of Iraq' is more persuasive in English. This could be the case as this effort was led by the US and UK, so writers in English would have more access to primary sources. Also more persuasive in English is the `2006 Kodori crisis'. This occurred in a separatist region of Georgia, and was alleged by Georgia officials to have been sponsored by agents of Russia. This is explored in more detail in English, while only briefly mentioned in Russian.

On the Russian side of the line (red), we have the `2005-2006 Russian-Ukraine gas dispute'. Interestingly, we also have the `First Battle of Brega', which was a 2011 conflict in the Libyan Civil War; neither Russia nor Anglosphere countries were directly involved. This example could warrant further study, into whether the Russian article contains more persuasion due to tastes of the particular author, or if the Russian media as a whole covered this war more.

\subsection{Verifying LLM's Consistent Understanding of English and Russian HLQs}
\label{sec:sanity}
Recall that for the persuasion detection task, merely giving the LLM the persuasion technique definitions resulted in the responses diverging from human labels. This advises us to also check whether the HLQs and prompts in English elicit similar behavior from an LLM as HLQs and prompts in Russian. After all, multilingual LLMs are largely English-centric; also most prior works advise to always use prompt instructions in English, even for inference in other languages~\citep{ahuja2023mega,shi2022language}.
Therefore, we perform a sanity check experiment, by considering settings RU and its translation RU2EN (and vice versa for and en2ru). As articles contain the same content, but just translated, we should expect their rankings to be similar; meanwhile, the rankings from the other language-authored articles should differ.

We use Rank Bias Overlap (RBO) to compare two ranked lists~\cite{webber2010rbo}. RBO is based on a simple probabilistic user mode, where higher scores (0 to 1) indicate more similar lists. These pairwise RBO scores are shown in Appendix Figure~\ref{fig:RBO}. The highest RBO is achieved between original and translated articles: RBO(ru, ru2en)$= 0.85$. In contrast, rankings differ greatly between the original articles: RBO(ru, en)$= 0.29$.

Therefore, we have shown that the HLQ-based approach to persuasive language detection is equally valid in either English or Russian. We also indirectly have shown that the translation process we used maintains the persuasive content of an original text.
To conclude this section, this set of experiments show the flexibility of our approach to uncovering cross-cultural differences in persuasion, from various angles.

\section{Conclusion}
\label{sec:conclusion}
Our study makes two contributions. First, we introduced the methodology of high-level questioning, in which we allow a LLM to author a large set of questions related to a subjective task, and then filter down to a target set of HLQs by aligning to human labels. Given this was highly effective for the binary persuasion task, we anticipate that the HLQ process can serve to improve LLM performance on other subjective tasks.

Second, we have made a large-scale inquiry into uncovering how Wikipedias in Russian and English differ in their perspectives. Our approach was to quantify levels of persuasive content used across different language versions of a subject. Our evaluation framework utilized several simple metrics to make deep insights into various questions: which subjects are most meaningful to authors in English and/or Russian? For which subjects are cross-lingual disagreements in persuasion highest? Future work can extend our analytical setup to additional cultural pairs.

Our work takes several preliminary steps towards using LLMs to enable large-scale cross-lingual insights. While cross-cultural differences exist, we are excited by the possibilities of multilingual LLMs, to facilitate better understanding across geographic and linguistic borders.

\section*{Limitations}

The main limitation of our work is that we collected the HLQs with respect to a labeled dataset (SemEval), and then applied it to an unlabeled dataset in a different domain (Wikipedia). Given the challenges of a domain adaptation setting, it would be ideal to have some labeled data in the target domain. However this was infeasible due to the size of our dataset (88k articles), and the extensive time and effort required to obtain labeled data that annotators agree on. We therefore proposed the series of experiments, based on the PF metrics, and found that the findings roughly matched our intuitions on subjects and cultural analysis. We anticipate followup work, such as the next iterations of SemEval, can further address the issue of requiring more labeled data for precision.

We also performed only a limited analysis of specific topics from Wikipedia, such as in \S\ref{sec:rank_persuasion}, \S\ref{sec:disagreement}. Followup studies should both consider more examples, and further investigate these examples with respect to the larger geopolitical landscapes of both Russian and English-speaking societies.

For ethical considerations, we used LLMs for the experiments throughout our work, and processed a high volume of text (88k Wikipedia articles). LLMs are known to use a large amount of compute. However, we tried to be efficient with our use of prompting LLMs; for example, our approach using 12 HLQs in 1 prompt, instead of 23 for each persuasion technique.

\bibliography{acl_latex}

\appendix
\begin{table*}[t!]
\small
\centering
\begin{tabular}{@{}lll|lll@{}}
\toprule
Persuasion Technique & \begin{tabular}[c]{@{}l@{}}Gold\\ Count\end{tabular} &  \begin{tabular}[c]{@{}l@{}}Baseline\\ Count\end{tabular} & Persuasion Technique & \begin{tabular}[c]{@{}l@{}}Gold\\ Count\end{tabular} & \begin{tabular}[c]{@{}l@{}}Baseline\\ Count\end{tabular} \\ \midrule
\textit{None} & 6945 & 1450 & Conversation Killer & 115 & 120\\
Loaded Language & 2277 & 2484 & Red Herring & 63  &  101\\ 
Name Calling-Labeling & 1226  &  1871 & Guilt by Association & 63  & 339\\ 
Doubt & 703  &  2824 & Appeal to Popularity & 48  & 478\\
Repetition & 684  &  407 & Appeal to Hypocrisy & 45  & 104\\
\begin{tabular}[c]{@{}l@{}}Exaggeration-\\ Minimisation\end{tabular} & 576 & 1571 & \begin{tabular}[c]{@{}l@{}}Obfuscation-Vagueness-\\ Confusion\end{tabular} & 30 & 482 \\
Appeal to Fear-Prejudice & 442  &  2260 & Straw Man & 24  & 19\\ 
Flag Waving & 376  &  46 & Whataboutism & 18  & 179\\ 
Causal Oversimplification & 236  &  848 & Appeal to Values & 0  & 1938\\ 
False Dilemma & 180 & 307 & \begin{tabular}[c]{@{}l@{}}Consequential \\ Oversimplification\end{tabular}    & 0 & 361 \\
Slogans & 180 & 124 & Appeal to Time & 0 & 577 \\
Appeal to Authority & 179  &  6286 & Questioning the Reputation & 0  & 483\\  \midrule
Total (all 23 excluding \textit{None}) & 7465 & \multicolumn{2}{l}{24209}  &  &  \\ \bottomrule
\end{tabular}
\caption{Raw counts of each persuasion technique for the SemEval English split~\citep{piskorski2023semeval}, gold vs GPT-4 baseline.}
\label{tab:pt_counts}
\end{table*}

\begin{table*}[t!]
\centering
\small
\begin{tabularx}{\linewidth}{ccX}
\hline
\textbf{QID} & \textbf{AN, FR} & \textbf{Question} \\
\hline
Q20  & 0, 3 & Does the author employ loaded language or emotionally-charged words? \newline \textcolor{blue}{\foreignlanguage{russian}{Использует ли автор насыщенный язык или эмоционально окрашенные слова?}}  \\
\hline
Q88  & 12, 7 & Is there any use of expressive language or  imagery that could influence the reader's feelings? \newline \textcolor{blue}{\foreignlanguage{russian}{Есть ли использование выразительного языка или образности, которые могут повлиять на чувства читателя?}} \\
\hline
Q92  & 9, 5  & Does the text make use of positive or  negative connotations to sway the reader's opinion? \newline \textcolor{blue}{\foreignlanguage{russian}{Использует ли текст позитивные или негативные коннотации для влияния на мнение читателя?}} \\
\hline
Q210 & 6, 9  & Does the text contain words or phrases that evoke strong emotions?  \newline \textcolor{blue}{\foreignlanguage{russian}{ Содержит ли текст слова или фразы, вызывающие сильные эмоции?}} \\
\hline
Q211 & 3, 2  & Are there words or phrases in the text  that are intended to manipulate the reader's feelings? \newline \textcolor{blue}{\foreignlanguage{russian}{Есть ли в тексте слова или выражения, предназначенные для манипулирования чувствами читателя?}} \\
\hline
Q212 & 1, 1  & Can you identify any instances where emotionally  charged language is used to support a claim? \newline \textcolor{blue}{\foreignlanguage{russian}{Можете ли вы указать случаи использования эмоционально окрашенных слов для поддержки утверждения?}}\\
\hline
Q213 & 8, 0  & Are there parts in the text where the language is  used to influence the reader's opinion or decision? \newline \textcolor{blue}{\foreignlanguage{russian}{Есть ли в тексте места, где язык используется для воздействия на мнение или решение читателя?}} \\
\hline
Q215 & 2, 31 & Does the text use language that is intended  to provoke a particular reaction from the reader? \newline \textcolor{blue}{\foreignlanguage{russian}{Использует ли текст язык, предназначенный для вызывания определенной реакции читателя?}} \\
\hline
Q216 & 5, 19 & Can you find any instances where the language used is not neutral or objective?\newline \textcolor{blue}{\foreignlanguage{russian}{ Можете ли вы найти случаи, когда используемый язык не нейтрален или объективен?}}  \\
\hline
Q217 & 7, 12 & Does the text use language that is intended to sway the reader's viewpoint?\newline \textcolor{blue}{\foreignlanguage{russian}{ Использует ли текст язык, предназначенный для влияния на точку зрения читателя?}}  \\
\hline
Q258 & 4, 4 & Are there any emotionally charged words or phrases used in the text?\newline \textcolor{blue}{\foreignlanguage{russian}{ Есть ли в тексте эмоционально окрашенные слова или выражения?}}  \\
\hline
Q295 & 20, 6 & Can you identify any instances where the text may be using hyperbole or understatement? \newline \textcolor{blue}{\foreignlanguage{russian}{Можете ли вы указать случаи, когда в тексте возможно использование гиперболы или преуменьшения?}} \\
\hline
\end{tabularx}
\caption{The 12 HLQs selected, with English in black and Russian in \textcolor{blue}{blue}. The second column shows the feature importance ranking by ANOVA (AN) and Random Forest (RF). In terms of persuasive techniques, we observe that 10 pertain to `Loaded Language', 1 (Q258) pertains to `None', and 1 (Q295) pertains to `Exaggeration or Minimization'. This reflects the overrepresentation of ``Loaded Language'' in SemEval (47\% of technique labels).} 
\label{tab:hlqs}
\end{table*}

\begin{figure} [t!]
\centering
\includegraphics[width=.975\linewidth]{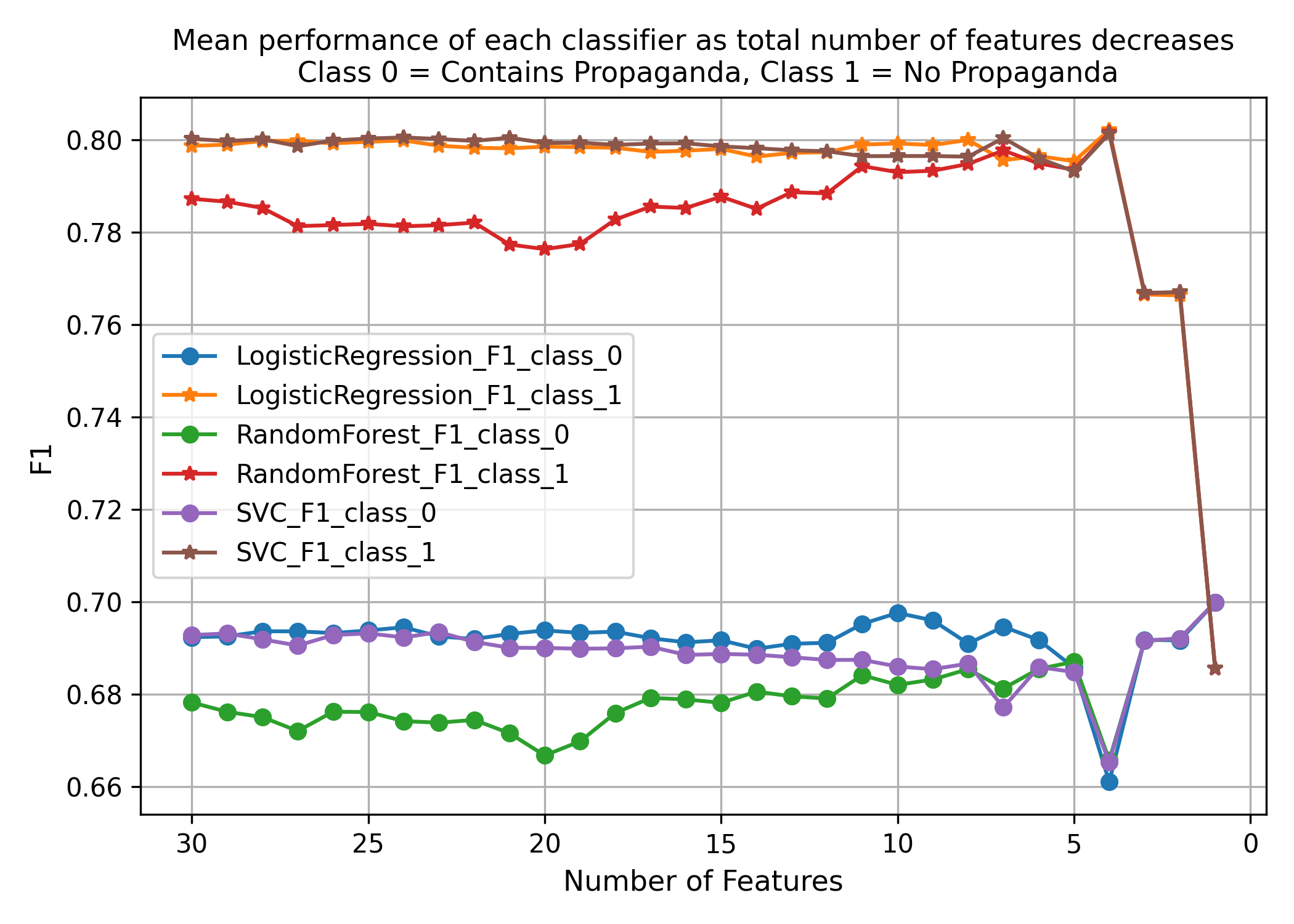}
\caption{ 
The effectiveness of the classifiers after each feature reduction using ANOVA. F1 performance is relatively stable across metrics from 30 to 8 features, and declines afterwards.}
\label{fig:feature_selection} 
\end{figure} 

\begin{table}[t!]
\centering
\small
\begin{tabularx}{\linewidth}{ccX}
\toprule
\textbf{Sent idx} & \textbf{QID} & \textbf{Specific Text Instances Identified} \\ \midrule
2 & Q20 & engulfed, rapidly destroyed, tragedy, repeatedly complained, ... \\ 
2 & Q88 & fire engulfed, rapidly destroyed, tragedy, funding cuts, ... \\ \midrule
 & \textit{PTS} & fire engulfed, rapidly destroyed, tragedy, repeatedly complained, funding cuts \\ \midrule
3 & Q20 & incalculable, outraged, cultural tragedy, lobotomy \\
3 & Q88 & fire, loss, outraged, tragedy, destroyed, ruins, threat, ...\\ \midrule
 & \textit{PTS} & incalculable, outraged, cultural tragedy, lobotomy, fire, loss, destroyed, ruins, threat \\ \bottomrule
\end{tabularx}
\caption{Sample Model responses (ru2en), on `Fire at the National Museum of Brazil' (WikiID: Q56441760). `PTS' is the deduplicated persuasive text set combining all 12 HLQs.}
\label{tab:questionResponseExample}
\end{table}

\begin{table}[ht!]
\centering
\small
\begin{tabularx}{\linewidth}{Xccc}
\toprule
\textbf{QID (Description)} & \textbf{\# Subjects} & \textbf{ru NPF} & \textbf{en NPF} \\ \midrule
Q180684 (Disagreement Situation) & 65 & 0.303 & 0.326 \\
Q47461344 (Written Work) & 53 & 0.301 & 0.305 \\
Q178561 (Part of War) & 68 & 0.304 & 0.284 \\
Q7278 (Org Influences Gov) & 138 & 0.247 & 0.296 \\
Q43229 (Social Entity) & 122 & 0.229 & 0.257 \\
... & ... & ... & ... \\
Q23038290 (Fossil Taxon) & 52 & 0.044 & 0.045 \\
Q15056993 (Aircraft) & 153 & 0.05 & 0.035 \\
Q786820 (Automaker) & 52 & 0.025 & 0.054\\
Q2198484 (Admin Entity) & 132 & 0.038 & 0.037 \\ 
Q14795564 (Date Calculator) & 217 & 0.036 & 0 \\ \bottomrule
\end{tabularx}
\caption{Top 5 and bottom 5 topics (Wikidata P31 \texttt{instance of}) by persuasive content. This is sorted by NPF en, but as shown, NPF en and NPF ru are mostly close over topics.}
\label{tab:top5_P31_Nodes}
\end{table}

\begin{figure} [t!]
    \centering
   \includegraphics[width=\linewidth]{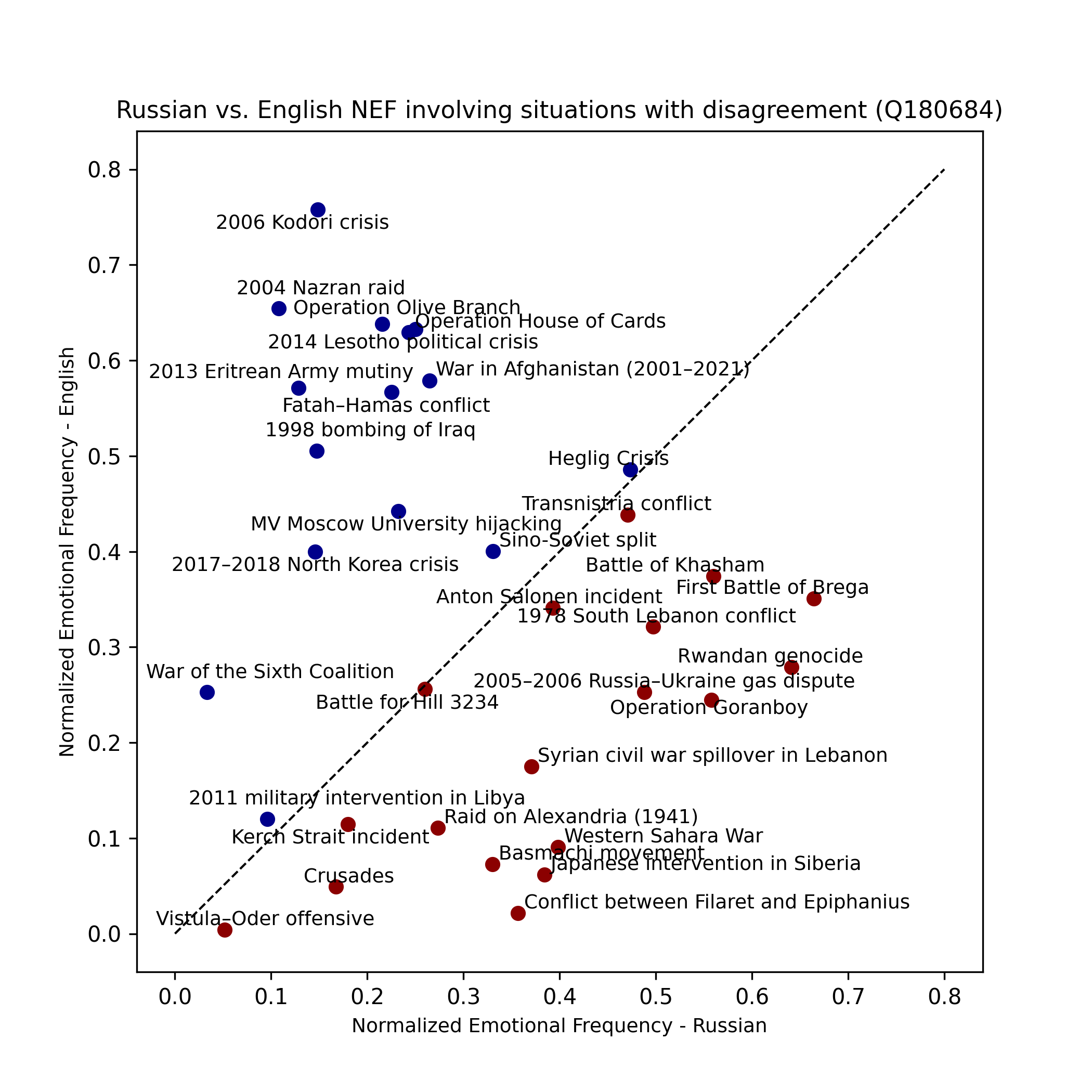}
   \caption{A scatterplot where the x and y positions represent the NPF values of Russian and English articles, respectively. The dashed line indicates equal NPF, i.e. the subjects where English and Russian has similar levels of emotional content. The further a point is from this line, the further the paired articles are in their use of persuasive content. }
   \label{fig:disagree}
\end{figure} 

\section{Discussion}
\subsection{Motivating high-level questioning}
\label{sec:why_high}
Let us consider the typical fixes one can take when an LLM underperforms given some prompt. First, more detailed instructions can be written. 
For this task, a human would have to expend efforts for all 23 techniques. Furthermore, longer instructions would great increase inference costs.

Second, we can include more few-shot exemplars. In the baseline, we used a single, static exemplar. Suppose one wanted to use multiple, dynamic exemplars. Typical prompting techniques would, for each inference entry, randomly draw exemplars from a train split. Again, this would be challenging due to the 23 distinct techniques, which have different class priors. Therefore, we are motivated to find an approach which can use the LLM's own intuitions.

\section{Additional Experiments}
\subsection{Normalized persuasion frequency (NPF)}
\label{sec:npf}
We use a normalized version of PF for several experiments. This normalizes all PF scores across all authors (either Russian or English) between 0 and 1. To quickly illustrate, suppose the max PF is 0.6, and the min is 0.05. NPF would draw the max towards 1, and the minimum towards 0. The raw max and min PF could differ between English and Russian, but after normalization, the max and min PF would be about the same.

We provide pseudocode for calulating NPF:
\begin{verbatim}
author1_pf = calc_pf(
  author1_article_length_list,
  author1_pc_list)
author2_pf = calc_pf(
  author2_article_length_list,
  author2_pc_list)
# Concatenate PF arrays from both authors  
all_pf = author1_pf + author2_pf
# Scale all PF values within the range [0, 1]
npf = normalize_scores(all_ef)
\end{verbatim}

Where \texttt{calc\_pf} returns \texttt{author\_pc\_list[i] / author\_article\_length\_list[i]} (for $i = 0, 1, ... n-1$), and $n$ is the number of articles.

\paragraph{Why normalize?}
Recall that Wikipedia guidelines specify a NPOV. If we assume that different individual authors aim for the same NPOV, then we can normalize PF scores for one language. Suppose the NPOV for Russian differs from English. Then, we can ``normalize'' out the NPOV by taking all article's and their PF together. Putting them on a common scale makes comparing the relative emotional content between authors more meaningful. We do acknowledge that this normalization makes several assumptions and is simplistic.

\begin{figure} [t!]
\centering
   \includegraphics[width=1\linewidth]{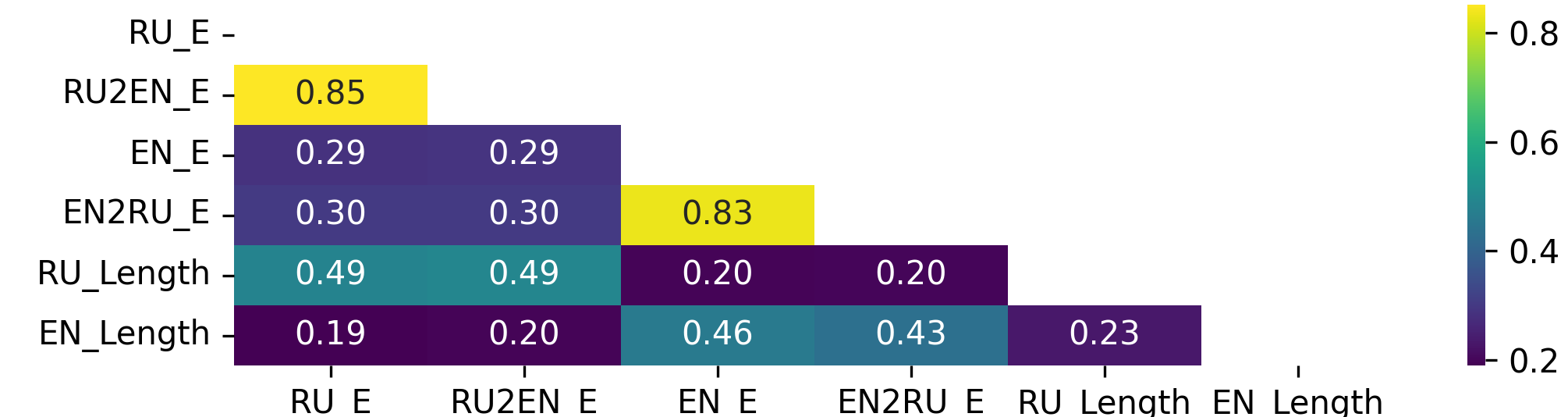}
   \caption{ 
Rank-biased overlap (RBO) scores, calculated over pairwise rankings. The rankings are the 4 language settings, as well as ru\_length and en\_length, which are $wc(\text{article})$. The label `\_E` refers to $PC$. }
   \label{fig:RBO} 
\end{figure}

\section{Identify-then-Extract with Llama-2}

\label{sec:llama}

\begin{table}[ht!]
\centering
\small
\begin{tabularx}{\linewidth}{ccX}
\toprule
\textbf{Index} & \textbf{QID} & \textbf{Specific Text Instances Identified} \\ \midrule
2 & Q20 &  \textcolor{blue}{negligence, \textbf{tragedy}, could have been avoided} \\ 
2 & Q88 & \textcolor{blue}{negligence, \textbf{tragedy}, could have been avoided}\\ \midrule
3 & Q20 & \textcolor{blue}{cultural \textbf{tragedy}, "incalculable" loss, \textbf{lobotomy} of Brazilian memory, ...} \\
3 & Q88 & \textcolor{blue}{cultural \textbf{tragedy}, "incalculable" loss, \textbf{lobotomy} of Brazilian memory, ...} \\ \midrule
\end{tabularx}
\caption{Llama responses to two questions for the article ``Fire at the National Museum of Brazil''. Text is given in  \textcolor{blue}{blue}, so as to compare to Table~\ref{tab:questionResponseExample}, with GPT-4 responses in black.}
\label{tab:questionResponseExampleWLlama}
\end{table}

For identify-then-extract with HLQs, we also ran a small study with Llama-2.\footnote{\url{https://huggingface.co/meta-llama/Llama-2-13b-chat-hf}} 
The rationale is that the decomposition of the harder task may enable smaller LLM's to perform reasonably. We do expect some performance drop, given the order of magnitude difference in size -- 13B vs >1T for GPT-4. Furthermore, given the closed-source nature of GPT-4, a locally-run, open-source model allows for more direct insights and analysis, especially for future work.

We report results for RU2EN, but have run the steps for all 4 settings. For ease of analysis and our computational budget, we restrict study to a 217 article subset of the original 22,046.

We found that several techniques were required to get Llama to adhere to the expected output format: more \textit{few-shot} examples, and \textit{pre-generating} the starting tokens of a response. These are explained ahead. Overall, we found that Llama underperformed GPT-4 in two other aspects: too many false positives, and shorter phrase extraction.

\subsection{Identify}
With the one-shot prompt, Llama had many errors in instruction-following -- it gives short answer responses with additional discussion. While under 10\% of Llama's initial responses were parseable, we achieved 90\% parseable responses by applying two modifications: 3-shot prompts, and pre-generation. The 3-shot prompts were manually curated, and then we manually wrote the persuasion responses. We selected paragraphs from 3 diverse articles -- a political article with considerable persuasion (Augusto Pinochet), a scientific article with a few instances of persuasion (Cobalt), and a scientific article with no persuasion (Banana).

Second, pre-generation follows from the observation that responses should always be prefixed with \texttt{Q1:} -- GPT-4 nearly always does this, while Llama by default rarely does. This leads to the intuition that we can \textit{pre-generate} the proper \texttt{Q1:} prefix by concatenating it to the input. Afterwards, the model will continue generations in this modified distribution space; we found that with pre-generation and few-shot, instruction-following improves to >90\%.
We note that this prompt engineering technique is similar to, for example, pre-generating "Answer: " for QA tasks.

\paragraph{Step 1 error analysis} 
Depsite the correct output format, as for the actual task, Llama output has several issues compared to GPT-4: it mostly outputs True, has much higher confidence scores (most are 90-100), and gives answers out of order (e.g. \texttt{Q0...Q1...Q9...Q4...}).

\subsection{Extract}
As with the identify step, we used few-shot prompts and pre-generation to enable better instruction-following. For few-shot prompts, we use the same 3 paragraphs, and write our few-shot examples for all questions. For pre-generation, we set the prefix to be a single quotation mark \texttt{"}.

\paragraph{Step 2 error analysis} 
Table~\ref{tab:questionResponseExampleWLlama} shows Llama's responses for the same article as in Table~\ref{tab:questionResponseExample} with GPT-4. We see that GPT-4 is able to extract longer clauses, while Llama prefers to extract short phrases. Also, the persuasive text sets (PTS) from Llama are shorter than those of GPT-4. Therefore, while it is feasible to use other LLMs with our persuasion detection approach, we decided to focus our experimental results on GPT-4.

\section{Prompts Used}
\label{sec:prompts}
Here we provide text for the prompts, exactly as used for the various LLM interactions. Note that these prompts slightly differ from those shown in the main text figures, which were edited for brevity.

:\\
\begin{figure*}[t]
    \centering
    \small
    \begin{tcolorbox}[enhanced, width=\linewidth, boxrule=0.8mm, colback=gray!5, colframe=gray!60]
\texttt{System}: Your task is to assign PersuasionTech types and confidence scores to given text (if more than one semicolon separated). You have a background in public relations, political science, and international relations. Confidence has integer value 0-100 (100 being the highest confidence). PersuasionTech has 24 possible values, here is value (definition) for each:\\ 
1. Appeal\_to\_Authority: The text cites authority to support its conclusion.\\ 
2. Appeal\_to\_Popularity: The text supports its conclusion by citing popularity or majority support.\\ 
3. Appeal\_to\_Values: The text invokes widely shared values to support its message.\\ 
4. Appeal\_to\_Fear-Prejudice: The text uses fear or prejudice to reject or promote an idea.\\ 
5. Flag\_Waving: The text refers to patriotism or group allegiance to back its conclusion.\\ 
6. Causal\_Oversimplification: The text oversimplifies the cause(s) of a subject or issue.\\ 
7. False\_Dilemma-No\_Choice: The text implies only two options when there may be more.\\ 
8. Consequential\_Oversimplification: The text oversimplifies the consequences of accepting a proposition.\\ 
9. Straw\_Man: The text misrepresents someone's position, usually to make it easier to attack.\\ 
10. Red\_Herring: The text diverts attention from the main topic.\\ 
11. Whataboutism: The text meant to distract from topic, discredits an opponent by charging them with hypocrisy.\\ 
12. Slogans: The text uses a brief, catchy phrase to encapsulate its message.\\ 
13. Appeal\_to\_Time: The text suggests that the time is ripe for a certain action.\\ 
14. Conversation\_Killer: The text discourages critical thought or discussion. \\
15. Loaded\_Language: The text uses emotionally charged words or phrases to validate a claim.\\ 
16. Repetition: The text repeatedly reinforces the same idea.\\ 
17. Exaggeration-Minimisation: The text either downplays or exaggerates a subject.\\ 
18. Obfuscation-Vagueness-Confusion: The text is deliberately unclear, leaving room for varied interpretations.\\ 
19. Name\_Calling-Labeling: The text employs demeaning labels to sway sentiments.\\ 
20. Doubt: The text attempts to undermine credibility by questioning character or attributes.\\ 
21. Guilt\_by\_Association: The text discredits an entity by associating it with a negatively viewed group.\\ 
22. Appeal\_to\_Hypocrisy: The text accuses the target of hypocrisy, often to tarnish their reputation.\\ 
23. Questioning\_the\_Reputation: The text undermines the reputation of the target, as a means to discredit their argument.\\ 
24. None: The text appears unbiased and doesn't evidently employ persuasion techniques.
\vspace{.5em} \hrule \vspace{.5em} 
\texttt{User}: Ukraine's government is “openly neo-Nazi” and “pro-Nazi,” controlled by “little Nazis,” President Vladimir V. Putin of Russia says.
    \end{tcolorbox}
    \centering
    \caption{Baseline prompt for persuasion detection.}
    \label{fig:baseline_prompt}
\end{figure*}

\begin{figure*}[t]
    \centering
    \small
    \begin{tcolorbox}[enhanced, width=\linewidth, boxrule=0.8mm, colback=gray!5, colframe=gray!60]
    \texttt{System:} Given a task X, your goal is to come up with a list of questions Y. The list Y contains questions
    that break the task into simpler components. Questions in list Y should be binomial: True or False.
    Questions in list Y should be semicolon separated. Avoid questions that rephrase the task, but do
    not simplify it.
    \vspace{.5em} \hrule \vspace{.5em} 
    \texttt{User:} \{Task\}: \{Task Definition\}
    \end{tcolorbox}
    \centering
    \caption{Prompt to generate HLQs for a Technique (zero-shot).}
    \label{fig:hlq_prompt}
\end{figure*}


\begin{figure*}[t]
    \centering
    \small
    \begin{tcolorbox}[enhanced, width=\linewidth, boxrule=0.8mm, colback=gray!5, colframe=gray!60]
    \texttt{System:} Given a piece of text your goal is to answer each of the following questions as 'True', 'False', or 'N/A' (if question is not applicable) plus a confidence measure from 0-100.\\
 Questions: \{list of 12 HLQs\}
 \vspace{.5em} \hrule \vspace{.5em} 
\texttt{User}: Ukraine's government is “openly neo-Nazi” and “pro-Nazi,” controlled by “little Nazis,” President Vladimir V. Putin of Russia says. 
 \vspace{.5em} \hrule \vspace{.5em} 
\texttt{Agent}: Q1: True (conf:70); Q2: False (conf:30); Q3: N/A; ...
    \end{tcolorbox}
    \centering
    \caption{Prompt for Identify stage of persuasive language detection.}
    \label{fig:id_prompt}
\end{figure*}


\begin{figure*}[t]
    \centering
    \small
    \begin{tcolorbox}[enhanced, width=\linewidth, boxrule=0.8mm, colback=gray!5, colframe=gray!60]
\texttt{System:} Given a piece of text your are tasked with a question: {Question} Identify specific language instances separated by semicolons.
Questions: \{list of 12 questions\}. 
\vspace{.5em} \hrule \vspace{.5em} 
\texttt{User:} Ukraine's government is “openly neo-Nazi” and “pro-Nazi,” controlled by “little Nazis,” President Vladimir V. Putin of Russia says. 
\vspace{.5em} \hrule \vspace{.5em} 
\texttt{Agent:} ``openly neo-Nazi''; ``pro-Nazi''; ``little Nazis''

    \end{tcolorbox}
    \centering
    \caption{Prompt for Extract stage of persuasive language detection. }
    \label{fig:extract_prompt}
\end{figure*}

\begin{figure*}[t]
    \centering
    \small
    \begin{tcolorbox}[enhanced, width=\linewidth, boxrule=0.8mm, colback=gray!5, colframe=gray!60]
\texttt{System:} Your task is to translate into English the given Russian text.

    \end{tcolorbox}
    \centering
    \caption{Prompt to translate English to Russian (zero-shot).}
    \label{fig:en2ru_prompt}
\end{figure*}

\begin{figure*}[t]
    \centering
    \small
    \begin{tcolorbox}[enhanced, width=\linewidth, boxrule=0.8mm, colback=gray!5, colframe=gray!60]
\texttt{System:}\otherlanguage{russian}{Ваша задача - перевести на русский язык данный английский текст.}

    \end{tcolorbox}
    \centering
    \caption{Prompt to translate Russian to English (zero-shot).}
    \label{fig:ru2en_prompt}
\end{figure*}

\end{document}